\documentclass{article}

    \PassOptionsToPackage{numbers, compress}{natbib}

\usepackage[final]{neurips_2021}

\usepackage[utf8]{inputenc} 
\usepackage[T1]{fontenc}    
\usepackage{hyperref}       
\usepackage{url}            
\usepackage{booktabs}       
\usepackage{amsfonts}       
\usepackage{nicefrac}       
\usepackage{microtype}      
\usepackage{xcolor}         
\usepackage[]{algorithm2e} 
\usepackage{amsmath} 
\usepackage{bm}
\usepackage{graphicx}
\usepackage[caption=false,font=normalsize,labelfont=sf,textfont=sf]{subfig}
 
\title{
A Collaborative Approach to the Analysis of the COVID-19 Response in Africa
}

\author{
  Sharon Okwako$^1$, Irene Wanyana$^2$, Alice Namale$^2$, Betty Kivumbi Nannyonga$^3$,\\
  \textbf{Sekou L. Remy$^1$, William Ogallo$^1$, Susan Kizito$^2$, Aisha Walcott-Bryant$^1$, Rhoda Wanyenze$^2$}\\
  $^1$IBM Research Africa\\
  Nairobi, Kenya\\
  $^2$Makerere University School of Public Health\\
  $^3$Makerere University School of Physical Sciences\\
  Kampala, Uganda\\
  \texttt{sharon@ke.ibm.com},
  \texttt{iwanyana@musph.ac.ug}\\
  \texttt{anamale@musph.ac.ug},
  \texttt{bnk@math.mak.ac.ug}\\
  \texttt{sekou@ke.ibm.com},
  \texttt{william.ogallo@ibm.com}\\
  \texttt{kizitosusan@musph.ac.ug},
  \texttt{awalcott@ke.ibm.com}\\
  \texttt{rwanyenze@musph.ac.ug}
  }

\begin{document}

\maketitle

\begin{abstract}
The COVID-19 crisis has emphasized the need for scientific methods such as machine learning to speed up the discovery of solutions to the pandemic. 
Harnessing machine learning techniques requires quality data, skilled personnel and advanced compute infrastructure.  
In Africa, however, machine learning competencies and compute infrastructures are limited. 
This paper demonstrates a cross-border collaborative capacity building approach to the application of machine learning techniques in discovering answers to COVID-19 questions.
\end{abstract}

\section{Introduction}
COVID-19 is a global challenge affecting lives worldwide \cite{WHOEU}.
According to the World Health Organization, as of $14^{th}$ September 2021, there were 225,024,781 confirmed cases of COVID-19 globally, including 4,636,153 deaths \cite{WHODASHBOARD}. 
In Africa, there are currently 5,757,213 reported cases of COVID-19 and 140,002 deaths as of $14^{th}$ September 2021 \cite{WHOAFRO}.
Given such high mortality and morbidity rates, countries require appropriate preparedness and response strategies to control disease spread and limit deaths \cite{piroth2021comparison}. 
There is also a need for timely and informed decision-making in implementing interventions and allocating resources, especially in low and middle-income countries with heavily burdened and fragile health systems \cite{makoni2020covid}. 

With the rapidly evolving COVID-19 pandemic and the urgent need to combat the disease, data aggregation, and analysis are critical for timely, effective, and efficient decision-making.
Unfortunately, the exponential increase in COVID-19 related data has rendered traditional data analysis techniques inefficient \cite{akinnagbe2018prospects}, thus requiring automated tools to extract hidden insights from massive data within the shortest time possible. 
Machine learning and artificial intelligence have been utilized to offer innovative solutions to public health concerns \cite{kushwaha2020significant}. 
They provide tools to support disease intervention planning, especially during a pandemic like COVID-19 \cite{blumenstock2020machine}. 
Several researchers have developed machine learning, and artificial intelligence techniques to aid the healthcare sector and policymakers forecast COVID-19 disease trends and diagnose and treat COVID-19 \cite{huang2021data,rashed2021one}.

In Africa, however, human resource and infrastructural capacities for machine learning and artificial intelligence are still very minimal. 
Furthermore, the increasing amount of health-related data is not equivalent to the available resources and techniques for its utilization \cite{akinnagbe2018prospects}. 

This project aimed to assess the COVID-19 pandemic responses in Africa and to build skills for machine learning within University Institutions in Africa. 
The institutions included Makerere University School of Public Health (Uganda) which was the lead university, the University of Ibadan in Nigeria, Cheikh Anta Diop University (Senegal), and the Kinshasa School of Public Health (Democratic Republic of Congo) in collaboration with IBM-Research Africa. 
These universities have a close working relationships with the Ministries of Health stakeholders in their respective countries and would  provide insights generated from the project to policy makers for decision-making. 

\section{Methodology}
\subsection{Research and Technical Collaboration}
The overarching research goal of this collaboration is to assess the impact of and response to the COVID-19 pandemic in East, Central, and West Africa. 
To this end, the collaboration is anchored on a multimodal program of research with several objectives such as 
(1) to document the government policies and response strategies to COVID-19, 
(2) to evaluate the effect of the response strategies (e.g. non-pharmaceutical intervention (NPI) measures such as lockdowns, mask-wearing, and social distancing) on the control of COVID-19, 
(3) to evaluate the effect of COVID-19 and related interventions on essential non-COVID care, 
(4) to evaluate the preparedness of health systems in handing COVID-19 and related interventions, and 
(5) to develop context-relevant strategies that inform policy and decision-making. 
To date, the collaboration has conducted the studies in the Democratic Republic of Congo, Nigeria, Senegal, and Uganda, with plans underway to including more partner countries.

\begin{figure}[ht]
\centering
\includegraphics[width=0.9\textwidth, height=9cm]{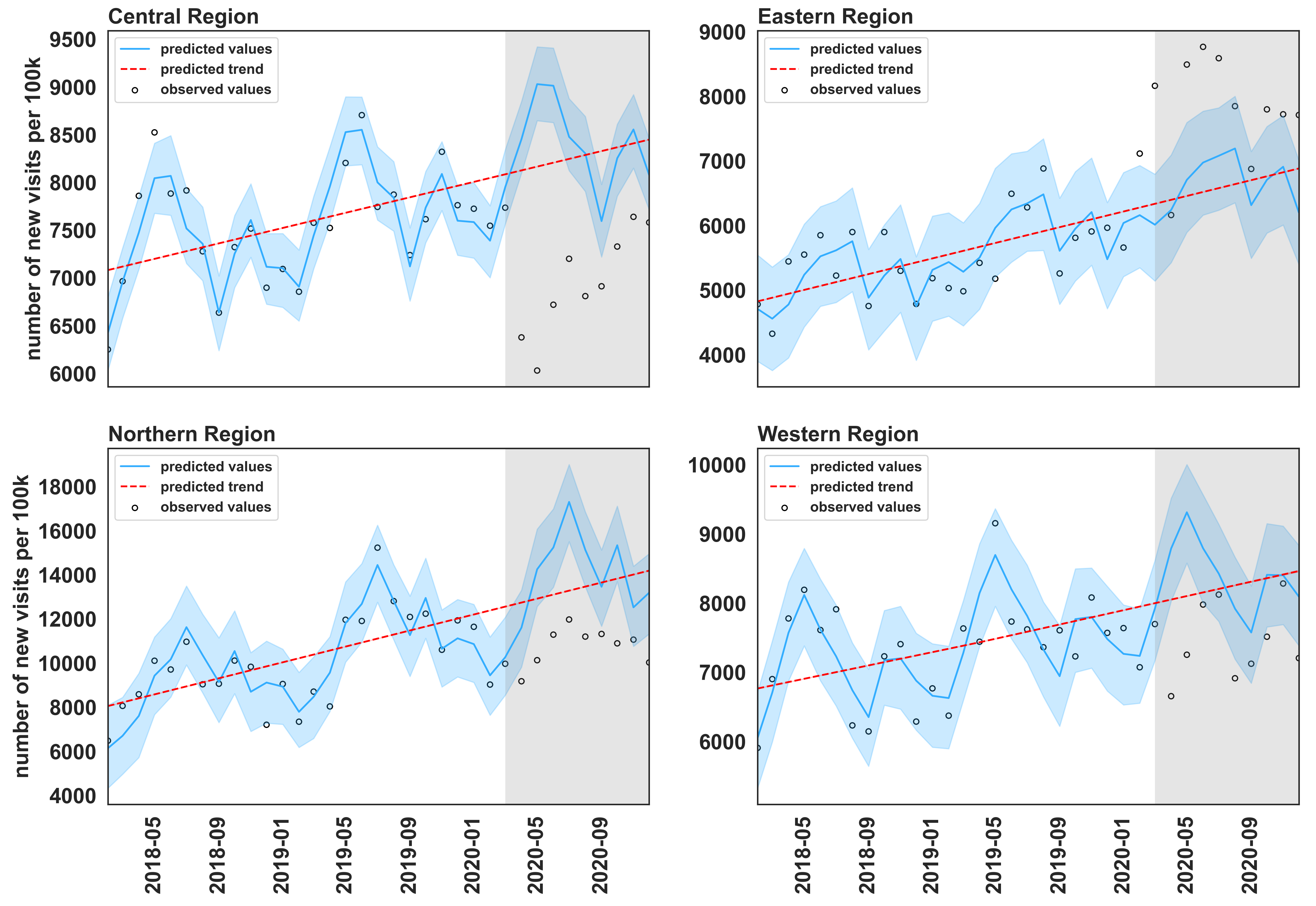}
\caption{Interrupted time series analyses suggest that during the Uganda COVID-19 pandemic period between March 2020 to December 2020, new clinic visits appear to have declined in the Northern, Central, and Western regions, and appear to have increased in the Eastern region}
\label{fig:1}
\end{figure}

To generate rich insights for decision-making, we combine traditional qualitative and quantitative research methods with state-of-the-art machine learning techniques for analyzing real-world evidence data. 
For example, to assess the impact of the COVID-19 pandemic on essential healthcare in Uganda, we used a mixed methods approach involving qualitative in-depth interviews (IDIs) and quantitative interrupted time series analyses (ITS). The qualitative IDIs aimed to describe the key informant perspectives on barriers to continuity of essential health service delivery. This investigation revealed several context-specific themes such as abandonment of critical policies, movement restrictions affecting health workers, suspension of outreach and support activities, and fear of fear of infection among patients. The quantitative ITS aimed to evaluate the impact of the introduction of COVID-19 NPIs on new clinic visits, diabetes clinic visits, and newborn hospital deliveries across the Central, Eastern, Northern, and Western regions of Uganda. As exemplified in Figure \ref{fig:1}, we found out that new clinic visits declined in the Northern, Central, and Western regions, and appear to have increased in the Eastern region. Furthermore, to amplify the value of COVID-19 population-based mathematical models (compartmental models), we applied model calibration based on artificial intelligence that incorporates information about the stringency of government responses to COVID-19 \cite{remy2020global}.

The core research team of our collaboration was comprised of an interdisciplinary cross-functional team of domain experts, research scientists, and software engineers, and business analysts working closely together and guided by agile values and principles. 
Throughout the collaboration, the team met weekly to discuss progress towards set goals and milestones, as well to discuss insights generated from the team's endeavors. 
Key decisions e.g., research questions to pursue, methodologies to use, and training and capacity building needs where made through collaborative consultations. 
To overcome the overheads of collaboration across different geographies, the team primarily met via online web conferencing applications.
\subsection{Training, Dissemination and Reuse}
\begin{figure}[ht]
\centering
\includegraphics[width=0.9\textwidth]{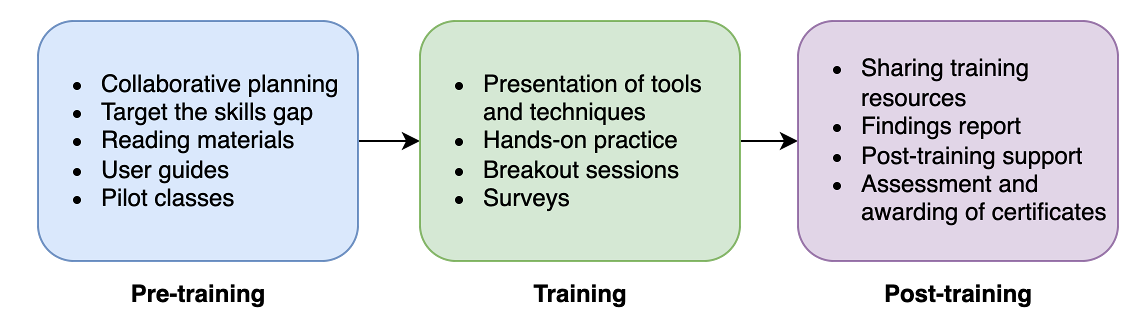}
\caption{Training and Capacity Building Model that was developed to guide the execution of the training sessions}
\label{fig:2}
\end{figure}

One key objective of this collaboration was establishing and strengthening a network/community of practice of universities focused on assessing the impact of public health emergencies of international concern in general and COVID-19 in particular for policy makers, based on existing networks housed by Makerere University School of Public Health (MakSPH). 
To this end, IBM Research-Africa
planned a number of virtual capacity building sessions to build participant's interest and capacity in data analysis and interpretation of results using available tools and resources. Some of the specific objectives of the sessions included teaching participants how to: 
1) Use of tools such as python for machine learning 
2) Prepare data and perform data analysis in python using examples from the MakSPH COVID-19 project
3) Interpret data analysis results and visualizations.  
The training content was hinged on the COVID-19 research work. COVID-19 exposed gaps in capacities in the developing countries to use available tools to support disease intervention planning. We used a collaborative approach to address the capacity gaps.
The virtual sessions offered an opportunity to reach a wider audience across the region.  
Figure \ref{fig:2} illustrates the training and capacity building model that was developed to guide the execution of the training and capacity building sessions. It comprised of the pre-training phase, training and post training phase. 
IBM Research-Africa worked closely with the MakSPH team to design and document the training and capacity building plan taking into account the needs of the participants. 
MakSPH assisted with recruiting participants from the respective institutions who had an interest in data science and experience with data analysis. 
Our target audience also included modelers, epidemiologists, policy makers, masters and PhD level students. 
Information about the training was broadcasted through different platforms, including study team meetings, emails, and social media groups (WhatsApp). 
The communication focused on study team members and students affiliated with the collaborating Universities. 
Those interested in participating were requested to register for the training. 
Interested participants were enrolled and added to a mailing list to enhance communication before and after the training. 
Given the diverse audience with different emerging technologies proficiency, we provided reading materials that helped participants familiarize themselves with the content and tools that would be used in the sessions. 

Our approach to the training consisted of two parts. 
The first part being lecture delivery on machine learning tools and techniques and the second part involved participants interacting with tools to conduct analyses. 
The topics covered in the sessions included interrupted time series analysis, descriptive analytics, correlation analysis and calibrating models to data. 
This required participants to have access to compute resources and a stable internet. 
The topics were chosen based on the analyses done by IBM Research Africa team to support the achievement of the MakSPH COVID-19 research objectives. 
Access to compute resources was provided free of charge through registration on IBM Digital Nation Africa (https://developer.ibm.com/digitalnation/africa/) where participants were directed for self-paced courses on emerging technologies such as artificial intelligence, cloud computing, data science and analytics among others. 
We had pilot classes with the MakSPH team where facilitators who are experts in machine learning and working on the research study walked the attendees through the content and the method of lecture delivery. 
This was particularly important in obtaining feedback on whether the planned session would meet the needs of the audience. 
We developed survey forms that were used in gauging participant's reactions to the sessions. 
The questions assessed the delivery of the presentations, the content and the facilitators on a Likert scale. 
The form also consisted of open-ended questions where participants were asked what they enjoyed most about the sessions, what they disliked and how the sessions could be improved. 





\section{Results}
\subsection{Training, Collaboration, and Reuse}
Through this collaboration, existing tools were extended with with additional features that fulfilled the needs of the users. 
Additionally, existing COVID-19 data sets in the available tools were updated to reflect the reality on the ground as partner institutions were more aware of the local contexts. 
These data sets are providing information which is crucial in contributing to decisions and mitigation strategies adopted for managing the pandemic. 
We conducted a number of virtual training and capacity building sessions on machine learning tools and techniques used for the quantitative assessment of the project. 
Participants from the 4 focus countries attended the sessions over an 18-week period covering 50+ hour lecture time. 
Each session ran for 90 minutes, and an average of 20 participants attended the training sessions.
The attrition rate was 39 percent. 
We were able to train the aforementioned target audience except policy makers. 
However, given the close working relationships between the universities and Ministry of Health policy makers, the participants can present the gained insights that would help improve policies.
Majority of participants did not have any background, knowledge or experience with machine learning techniques. 
Participants including those with minimal knowledge in machine learning demonstrated interest in gaining the skills. Given this background, an introduction to writing code in python and artificial intelligence was included in the teaching materials. This was not planned before.   
32 males and 17 females attended at least one training session revealing gender disparities in the machine learning space. 
Majority of the participants came from Anglophone countries. 

The facilitators presented the tools and analysis done for the research study which was followed by hands-on sessions where participants used the tools to generate or view the outputs. 
This was made possible through a cloud computing infrastructure, Watson studio, which allowed participants to access jupyter notebooks without requiring installation and configuration in their local machines. Participants logged into the platform, imported jupyter notebooks from github or through a shared link and were able to interact with the notebooks. 
The availability of this infrastructure made training easier. 
In one session, we utilized breakout sessions with the goal of ensuring that everyone could access the data and add the notebook in the compute resource provided. 
During the sessions, participants experienced unstable internet which was a hindrance to their effective learning and participation.
Some participants reported that they were not successful in loading the notebooks due to internet challenges. 
Some reported that they were not able to follow through the sessions.  
We had bilingual participants from Francophone countries but were not fluent in English and found the explanations too fast for them. 
Given that the tools were in English, these participants expressed their interest in having the tools translated to French.
For some tools, automatic google translation of web content helped the participants follow the discussions. 
All sessions were recorded and recordings shared with the participants for their reference. 
For the capacity building sessions on modeling, participants were required to demonstrate their ability to calibrate a model to data and 50 percent of participants successfully completed the assessment and obtained a certificate. 

The findings from the survey reveal that participants enjoyed the practical sessions the most. 
However, participants felt that they were not given enough time for hands-on practice and too much content was covered in little time. 
Participants expressed their interest in learning how to apply machine learning methods to alternative data sets apart from COVID-19 and the use of the codes in their daily work. 
On areas of improvement, participants stated that 
1) more time should be given to hands-on session 
2) recapping previously learnt concepts would help reinforce concepts 
3) facilitators should ensure that participants are on the same page as them (individual based follow-up) 
4) More use of break-out sessions for large groups to improve learning outcomes. 
The tools and codes used in the training sessions are open-source and publicly available. Some of the tools, codes and dataset can be accessed at https://ibm.github.io/wntrac/.   
This allows participants to continue using the available resources for further learning. 

The impact of these capacity building efforts cannot be understated. 
Several members in the consortium of universities have picked interest in learning more about machine learning and data science. 
More advanced members are utilizing the freely available resources such as codebase and tools and acquired knowledge to analyze existing data. 
The members have a better understanding of the role and impact of machine learning in generating valuable insights for solving complex problems. 

\section{Study Limitations}
We recognize that while the study results are valuable, several limitations exist in the study. 
First, the research study was conducted for Francophone and Anglophone countries. 
However, English was chosen as the language of communication and research excluding participants from Francophone countries who do not speak English or are not quite conversant with the language. 
Secondly, knowledge retention and application of skills takes time. 
Having a number of capacity building sessions may not be enough to sustain behaviour change.
Thirdly, our mode of engagement was entirely remotely which was not effective for members who did not have reliable internet. 
\section{Conclusion and Way Forward}
This paper describes our collaborative approach and experience in researching and developing machine learning competencies in Africa. 
We find that processes and tools play a critical role in improving effectiveness of capacity building efforts. 
The challenges we experienced in the process such as unstable internet and language barriers offers an opportunity for the development of innovative solutions to tackle them. 
There is need to develop tools that can be effectively used in poor internet connectivity or tools that do not require internet. 
The option of setting up environments in local machines can be explored further in a way that supports participants efficiently. 

Several initiatives have been put in place to address the skills gap in the continent. 
These initiatives include machine learning communities, courses, and also funding opportunities for machine learning research. 
However, from our experience, many are not aware that these initiatives or resources exist. 
There is therefore need to publicize these initiatives and more collaboration between industry stakeholders and academic institutions should be explored. 
On the other hand, the available materials are often written or produced in one language, therefore excluding people in the continent who only speak other languages. 
This offers a great opportunity for the creation of more linguistically diverse resources to support development of machine learning in the continent. 
This unique collaboration between private sector and universities in the region encourages the creation of such partnerships that allows flow of knowledge, skills and resources which will lead to improved research outputs. 
To ensure sustained interest and continued skills improvement and utility for country programs, there is need for further engagement in the form of communities which members having access to experts for continued guidance and mentorship.



\section{Acknowledgements}

We would like to thank Bill and Melinda Gates Foundation for funding the research collaboration that included training of partner members. 

\newpage

{\small \bibliography{main}}
\bibliographystyle{unsrt}

\end{document}